\documentclass[letterpaper, 10 pt, conference]{ieeeconf}  % Comment this line out if you need a4paper

\IEEEoverridecommandlockouts                              % This command is only needed if 
\usepackage{url}
\usepackage{cite}
\usepackage{graphicx}
\usepackage{amsmath,amssymb,amstext}
\usepackage{subcaption}

\usepackage{caption}
\usepackage{tabularx}
\usepackage{hyperref}
\hypersetup{
    pdfnewwindow=true,      % links in new window
    colorlinks=true,        % false: boxed links; true: colored links
    linkcolor=blue,         % color of internal links
    citecolor=green,        % color of links to bibliography
    filecolor=magenta,      % color of file links
    urlcolor=cyan           % color of external links
}

\usepackage{titlesec}
\titlespacing*{\section} {0pt}{0.75ex}{0.75ex}
\titlespacing*{\subsection} {0pt}{0.75ex}{0.75ex}

\title{\LARGE \bf
Learning Skateboarding for Humanoid Robots through Massively Parallel Reinforcement Learning
}
\author{William Thibault$^{1}$, Vidyasagar Rajendran$^{1}$, William Melek$^{1}$ and Katja Mombaur$^{1,2}$% <-this % stops a space
\thanks{$^{1}$William Thibault ({\tt\small wcthibau@uwaterloo.ca}), Vidyasagar Rajendran and William Melek are with the University of Waterloo, Waterloo ON N2L 3G1, Canada}%
\thanks{$^{2}$Katja Mombaur is with Karlsruhe Institute of Technology, 76131 Karlsruhe, Germany}
}

\author{William Thibault$^{1}$, Vidyasagar Rajendran$^{2}$, William Melek$^{1}$, Katja Mombaur$^{2,3}$
\thanks{$^{1}$William Thibault and William Melek are with Department of Mechanical and Mechatronics Engineering, University of Waterloo,
        Waterloo, ON N2L 3G1, Canada
        {\tt\small \{wcthibau, wmelek\}@uwaterloo.ca}}%
\thanks{$^{2}$Vidyasagar Rajendran and Katja Mombaur are with the Department of Systems Design Engineering, CERC Human-Centred Robotics and Machine Intelligence,
        University of Waterloo, Canada
        {\tt\small \{vrajendr, kmombaur\}@uwaterloo.ca}}%
\thanks{$^{3}$Katja Mombaur is also with the Institute for Anthropomatics and Robotics (IAR), Optimization and Biomechanics for Human-Centred Robotics, Karlsruhe Institute of Technology (KIT), Karlsruhe, Germany
        {\tt\small katja.mombaur@kit.edu}}%
}

\begin{document}
\maketitle
\thispagestyle{empty}
\pagestyle{empty}
%%%%%%%%%%%%%%%%%%%%%%%%%%%%%%%%%%%%%%%%%%%%%%%%%%%%%%%%%%%%%%%%%%%%%%%%%%%%%%%%
\begin{abstract}
Learning-based methods have proven useful at generating complex motions for robots, including humanoids. Reinforcement learning (RL) has been used to learn locomotion policies, some of which leverage a periodic reward formulation. This work extends the periodic reward formulation of locomotion to skateboarding for the REEM-C robot. Brax/MJX is used to implement the RL problem to achieve fast training. Initial results in simulation are presented with hardware experiments in progress. 
\end{abstract}
%%%%%%%%%%%%%%%%%%%%%%%%%%%%%%%%%%%%%%%%%%%%%%%%%%%%%%%%%%%%%%%%%%%%%%%%%%%%%%%
% \vspace{-0.2cm}
\section{INTRODUCTION}
Humanoid robots have been gaining popularity recently due to the advantage of having general-purpose robots that can operate in human-like environments to fill in labor shortages for undesirable tasks. For any application, the core skills required to interact and move about the environment involve manipulation and locomotion. Advances in machine learning have lead to the successful use of learning-based techniques, including reinforcement learning (RL), for learning these core skills. Many works have shown that bipedal walking can be learned for robust and stable walking on real robots making learning-based approaches appealing for generating complex motions for humanoids \cite{rodriguez2021,radosavovic2023,singh2023}.
Due to the periodic nature of walking, periodic reward formulations, as originally suggested by Siekmann et al. \cite{siekmann2021}, have been a popular way to learn bipedal walking as it allows for easily describing contact and flight phases of walking to produce a symmetrical gait. This periodic reward formulation proved crucial for humanoid robots that have large, heavy limbs like HRP-5P from Singh et al. 2022, 2023 \cite{singh2022,singh2023}. Recently, a periodic reward formulation that leveraged the Brax RL algorithms \cite{brax2021github} with MJX \cite{mujoco_mjx}, the XLA version of the MuJoCo simulator \cite{todorov2012}, was used to train the REEM-C humanoid robot to perform joystick walking in under 1 hour. Similarly to Issac Gym which has proven to be powerful for reinforcement learning problems like quadruped walking \cite{rudin2022}, Brax and MJX allow for massively parallel training with some benchmarks suggesting better training performance \cite{caluwaerts2023}. Periodic reward formulations allow for flexible description of repetitive gait cycles, leading to the possibility of other periodic gait motions  like skipping or hopping \cite{siekmann2021}. 
%
% \vspace{-0.225cm}
\begin{figure}[t]
    \centering
    {\includegraphics[width=0.4\linewidth, trim={0cm 0cm 0cm 0cm}, clip]{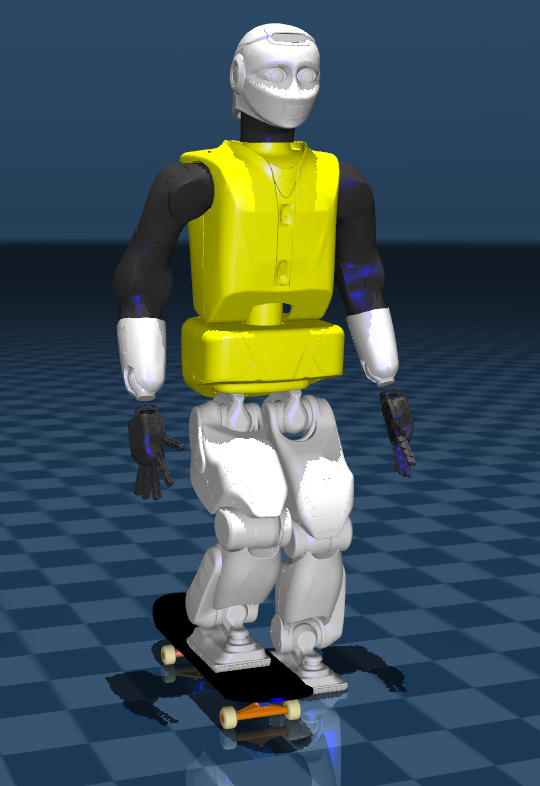}}
    % \hfill
    {\includegraphics[width=0.4\linewidth, trim={0cm 0cm 0cm 0cm}, clip]{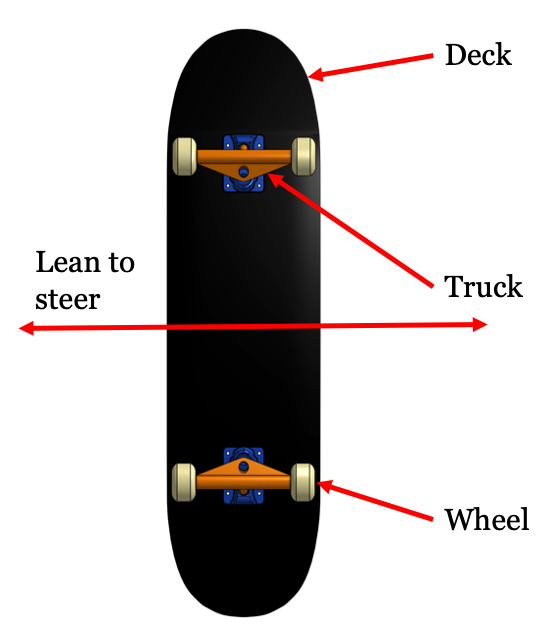}}
    \caption{REEM-C on skateboard (left), skateboard (right).}
    \label{fig:reemc_and_deck}
    \vspace{-0.8cm}
\end{figure}

Skateboarding is a complex bipedal task that involves using one foot to push on the ground while the other foot is on the skateboard to propel forward, while also balancing side to side to maintain a straight trajectory. Previously, the small hybrid bipedal, drone robot LEONARDO was shown to skateboard by using its propellers to push itself rather than using its feet while leaning side to side to turn \cite{Kim2021}. Another work demonstrated a skating like motion with HRP-2 using a skateboard-like device that does not involve lateral balance (a board with fixed wheels) and a kick board that has a handle \cite{takasugi2016}. However, skateboarding for a full size humanoid robot on a real skateboard has yet to be achieved. This work explores the possibility of leveraging RL with a periodic reward formulation for the cyclic motion of skateboarding on the REEM-C robot as an extension of the walking formulation of Thibault el. al \cite{thibault2024}.
\section{METHOD}
\subsection{REEM-C Robot and Simulation Environment}
The goal of learning joystick-style velocity skateboarding involves the actuated REEM-C robot and a passive skateboard. The REEM-C robot is a full-size humanoid robot with 30 degrees of freedom (DoF): two 6 DOF legs, two 7 DoF arms, a 2 DoF torso and a 2 DoF head. Similarly to Thibault et al. 2024 \cite{thibault2024}, only the 12 DoF for the legs are actuated with the other joints held fixed to simplify the learning problem. The initial position of the REEM-C on the skateboard and the model of the skateboard can be seen in Figure \ref{fig:reemc_and_deck} where the right foot is on the deck of the skateboard and the left is on the ground. The skateboard has four passive wheels that are attached to the board, known as the deck, by two passive trucks. The trucks are used to hold the wheels while attached to the board so that when leaning more towards one edge of the deck, the trucks flex to turn the wheels and in turn the board in the direction of the leaning. Both the REEM-C robot and skateboard are modelled with complete dynamics with floating bases in the MJX simulator. The contacts of the feet with the deck and ground are unilateral as are the contacts of the wheels with the ground. For theses contacts, increased friction is applied for the contact of the right foot with the deck to simulate the effect of the grip tape that is normally on the top surface of the deck. With respect to the trucks, a stiffness was added to the joint to replicate the spring-like effect of trucks to remain centered when there is little leaning, but turned when leaning.

\subsection{RL Problem Formulation}
This problem formulation is a direct extension of Thibault et al. \cite{thibault2024}, thereby leveraging the same action space, observation space and rewards with a few exceptions. With a periodic reward formulation, contact and flight phase rewards are used to describe the behaviour of the feet for a cyclic motion. In this case, a single foot repeated pushing motion is formulated where the right foot remains on the skateboard and the left foot pushes on the ground then lifts repeatedly. For this pushing motion the command velocity that the robot's center of mass and the skateboard deck will attempt to track is limited to the forward direction only: (x, y, $\psi$) = ([0.0, 1.0]m/s, 0.0 m/s, 0.0 rad/s). The observation space is augmented with information about the skateboard including the 3 dimensional deck position and the 6 dimensional deck linear and angular velocity. Given the periodic reward formulation, the indicator function for gait phases is defined with a 0.75 s double support phase and a 1.0 s single support phase with the left foot in swing and the right foot on the skateboard. The exact same reward formulation and weights are sufficient to learn the skateboarding problem; however, the following rewards are added for faster learning along with improved behaviour for the deck's motion and the right foot to deck contact:
\begin{itemize}
    \item Deck linear velocity tracking reward: This reward tracks the target linear velocity for the deck.\\
    $r_{1} = e^{- || v_{xy,input}-v_{xy,deck} ||_2^2/\sigma}$\\
    where $v_{xy,input}$ is the XY command velocity, $v_{xy,deck}$ is the XY deck velocity and $\sigma$ is a scaling factor.
    \item Deck angular velocity tracking reward: This reward tracks the target angular velocity for the deck.\\
    $r_{2} = e^{-(\omega_{z,input}-\omega_{z,deck})^2/\sigma}$\\
    where $\omega_{z,input}$ is the command angular velocity, $\omega_{z,deck}$ is the deck velocity  and $\sigma$ is a scaling factor.
     \item Deck foot world velocity reward: This is an additional reward for the right foot to move while in contact.\\
    $r_{3} = ||v_{right}||_2^2$\\
    where $v_{right}$ is the velocity of the right foot.
    \item Foot slip with respect to deck penalty: This reward penalizes translation of the right foot on the deck to keep the foot centered on the deck.\\
    $r_{4} = ||v_{xy,deck} - v_{xy,right}||_2^2$\\
    where $v_{xy,deck}$ is the XY velocity of the deck and $v_{xy,right}$ is the XY velocity of the right foot.
    \item Foot rotation with respect to deck penalty: This reward penalizes roll and pitch velocities of the right foot on the deck to keep the foot contact flat.\\
    $r_{5} = ||\omega_{xy,deck} - \omega_{xy,right}||_2^2$\\
    where $\omega_{xy,deck}$ is the roll and pitch angular velocities of the deck and $\omega_{xy,right}$ is the roll and pitch angular velocities of the right foot.
\end{itemize}
%%%%%%%%%%%%%%%%%%%%%%%%%%%%%%%%%%%%%%%%%%%%%%%%%%%%%%%%%%%%%%%%%%%%%%%%%%%%%%%%
\section{RESULTS}
\begin{figure}[h]
    \vspace{-0.225cm}
    \centering
    {\includegraphics[width=0.22\linewidth, trim={2cm 0cm 2cm 0cm}, clip]{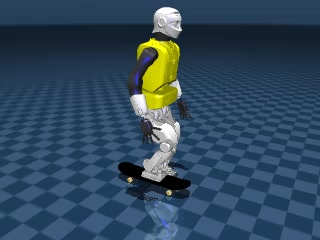}}
    % \hfill
    {\includegraphics[width=0.22\linewidth, trim={2cm 0cm 2cm 0cm}, clip]{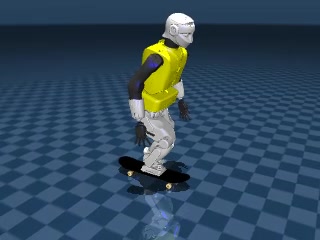}}
    % \hfill
    {\includegraphics[width=0.22\linewidth, trim={2cm 0cm 2cm 0cm}, clip]{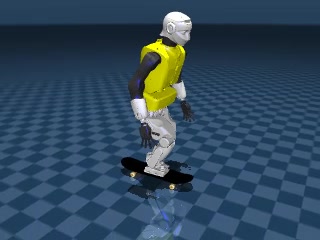}}
    % \hfill
    {\includegraphics[width=0.22\linewidth, trim={2cm 0cm 2cm 0cm}, clip]{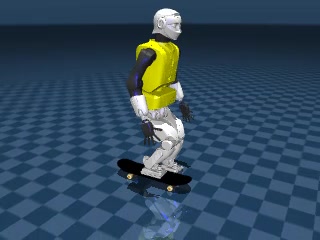}}
    \\
    \vspace{3px}
    {\includegraphics[width=0.22\linewidth, trim={2cm 0cm 2cm 0cm}, clip]{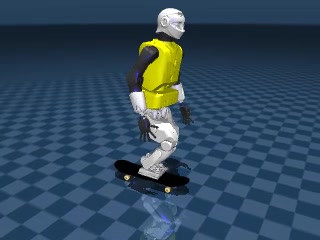}}
    % \hfill
    {\includegraphics[width=0.22\linewidth, trim={2cm 0cm 2cm 0cm}, clip]{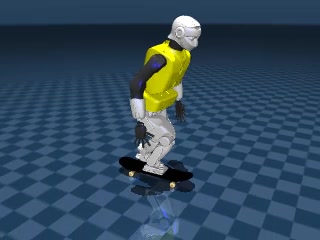}}
    % \hfill
    {\includegraphics[width=0.22\linewidth, trim={2cm 0cm 2cm 0cm}, clip]{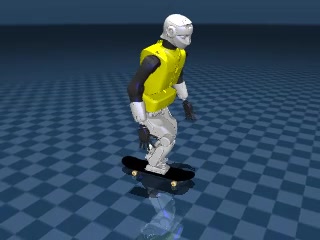}}
    % \hfill
    {\includegraphics[width=0.22\linewidth, trim={2cm 0cm 2cm 0cm}, clip]{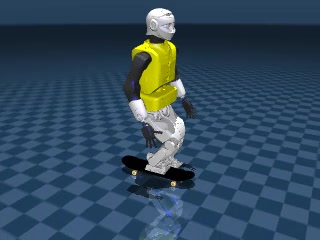}}
    \caption{REEM-C skateboarding forward at 0.4 m/s}
    \label{fig:sim}
    \vspace{-0.5cm}
\end{figure}

The RL training was set up with Brax and MJX for highly parallel training with 8192 parallel environments for 200,000,000 training steps using PPO similarly to Thibault et al. \cite{thibault2024}. This RL training pipeline was selected as it allows for fast wall clock training times, while leveraging the simulation behaviour of MuJoCo. A skateboarding motion sequence can be seen in Figure \ref{fig:sim}.

In the simulation results, it can be seen that the left foot is used to push along the ground until the foot nears the back wheels of the skateboard then it is lifted and placed near the front wheels of the skateboard to perform the next push. The right foot remains flat and centered on the deck during the pushing motion. An interesting emergent behaviour occurs with the unactuated upper body during the push phase of the motion. The upper body leans forward to maintain balance while the robot pushes itself forward similar to the way a human would skateboard. The overall motion achieved is smooth and balanced with little turning.
%
%%%%%%%%%%%%%%%%%%%%%%%%%%%%%%%%%%%%%%%%%%%%%%%%%%%%%%%%%%%%%%%%%%%%%%%%%%%%%%%%
\section{CONCLUSION}
In this work, initial results for a humanoid skateboarding RL policy based on a periodic reward formulation was shown for the REEM-C robot. This work extends Thibault et al. \cite{thibault2024} through a new gait cycle and added rewards for improved skateboarding performance while leveraging the fast, massively parallel training approach with Brax/MJX. On going work involves transferring the learned skateboarding abilities to the real REEM-C robot. Additionally, the skateboarding abilities are being extend to include a glide phase with both feet on the skateboard and the ability to turn.
%%%%%%%%%%%%%%%%%%%%%%%%%%%%%%%%%%%%%%%%%%%%%%%%%%%%%%%%%%%%%%%%%%%%%%%%%%%%%%%%
\section{ACKNOWLEDGEMENT}
We acknowledge the support of the Natural Sciences and Engineering Research Council of Canada (NSERC), the Tri-Agency Canada Excellence Research Chair Program, the University of Waterloo, the Karlsruhe Institute of Technology and Hector Stiftung.
%%%%%%%%%%%%%%%%%%%%%%%%%%%%%%%%%%%%%%%%%%%%%%%%%%%%%%%%%%%%%%%%%%%%%%%%%%%%%%%%
% \bibliography{references.bib}
% \bibliographystyle{IEEtrans}
\bibliographystyle{./IEEEtran} % use IEEEtran.bst style
\bibliography{./IEEEabrv,./references}
\end{document}